\documentclass[12pt]{amsart}
\usepackage{amsbsy,amssymb,amsmath,amsthm,amscd,amsfonts,latexsym,amstext,delarray,
amsmath,graphicx,url} 
\usepackage{qtree}
\usepackage[margin=.8in]{geometry}
\usepackage{color}
\usepackage[all]{xy}

\numberwithin{equation}{section}

\def\Q{{\mathbb Q}}

\def\K{{\mathbb K}}

\def\cA{{\mathcal A}}

\def\cD{{\mathcal D}}
\def\cE{{\mathcal E}}

\def\cH{{\mathcal H}}
\def\cI{{\mathcal I}}

\def\cM{{\mathcal M}}

\def\cO{{\mathcal O}}
\def\cP{{\mathcal P}}

\def\cS{{\mathcal S}}

\def\cV{{\mathcal V}}

\def\bY{{\mathbb Y}}

\def\fB{{\mathfrak B}}
\def\fM{{\mathfrak M}}
\def\fT{{\mathfrak T}}
\def\fF{{\mathfrak F}}

\title{Old and New Minimalism: a Hopf algebra comparison}
\author{Matilde Marcolli, Robert C.~Berwick, Noam Chomsky}
\date{2023}
\address{Departments of Mathematics and of Computing and Mathematical Sciences, 
California Institute of Technology, Pasadena, CA 91125, USA}
\email{matilde@caltech.edu}
\address{Institute for Data, Systems, and Society,
Massachusetts Institute of Technology, 
Cambridge MA 02141, USA}
\email{berwick@csail.mit.edu}
\address{Department of Linguistics, 
University of Arizona, Tucson, AZ 85721, USA}
\email{noamchomsky@email.arizona.edu}
\email{chomsky@mit.edu}

\begin{document}
\maketitle

\begin{abstract}
In this paper we compare some old formulations of Minimalism, in
particular Stabler's computational minimalism, and Chomsky's new formulation
of Merge and Minimalism, from the point of view of their mathematical
description in terms of Hopf algebras. We show that the newer formulation
has a clear advantage purely in terms of the underlying mathematical structure.
More precisely,  in the case of 
Stabler's computational minimalism, External Merge can be described in terms of a
partially defined operated algebra with binary operation, while Internal Merge
determines a system of right-ideal coideals of the Loday-Ronco Hopf algebra 
and corresponding right-module coalgebra quotients. This mathematical structure
shows that Internal and External Merge have significantly different roles in the
old formulations of Minimalism, and they are more difficult to reconcile as facets
of a single algebraic operation, as desirable linguistically. On the other hand,
we show that the newer formulation of Minimalism naturally carries a Hopf
algebra structure where Internal and External Merge directly arise from the same
operation. We also compare, at the level of algebraic properties, the externalization 
model of the new Minimalism with proposals for assignments of planar embeddings 
based on heads of trees. 
\end{abstract}

\section{Introduction}

The {\em Minimalist Program} of generative linguistics, introduced by Chomsky
in the '90s, \cite{Chomsky95}, underwent a significant simplifying 
reformulation in the more recent work \cite{Chomsky13}, \cite{Chomsky17}, 
\cite{Chomsky19}, \cite{Chomsky21} (see also \cite{BerCh}, \cite{BerEp}, \cite{Komachi}).
A recent overview of the current formulation of the fundamental Merge operation of
the Minimalist Model is presented in the lectures \cite{Chomsky23}.

\smallskip

In our paper \cite{MCB} we showed that the new formulation of Merge has a very
natural mathematical description in terms of magmas and Hopf algebras. This
mathematical formulation makes it possible to derive several desirable linguistic
properties of Merge directly as consequences of the mathematical setting.
We also showed in \cite{MCB} that the mathematical formulation of Merge has the same
structure as the mathematical theory underlying fundamental
interactions in physics, such as the renormalization process of quantum
field theory and the recursive solution of equations of motion via combinatorial
Dyson-Schwinger equations. 
An analogous recursive generative process of hierarchies of 
graphs plays a crucial role in both cases, as Feynman graphs in the physical case
and as syntactic trees in the linguistic case.

\smallskip

In the present companion paper, we show how this same mathematical 
formalism based on Hopf algebras can be used to compare the new
formulation of Merge with older forms of the Minimalist Model. The
advantages of the new formulation can be stated directly in linguistic
terms, as discussed in \cite{Chomsky17}, \cite{Chomsky19}, \cite{Chomsky23},
for instance. What we argue here is that one can also see the advantage of
the New Minimalism directly in terms of the underlying mathematical structure.

\smallskip

More precisely, we first consider the case of Stabler's ``Computational 
Minimalism", \cite{Stabler}. This formulation 
is unsatisfactory from the linguistic perspective, and superseded by Chomsky's
more recent formulation of Minimalism. We take this as an example because among 
the older versions of Minimalism it is one that tends to be more widely known to mathematicians
(as well as to theoretical computer scientists),
through its relation to formal languages. Indeed, mathematicians
familiar with the theory of formal languages are usually aware of the fact that Stabler's formulation
of Minimalism is describable in terms of a class of {\em minimalist grammars} (MG), 
which are equivalent to {\em multiple context free grammars}, a class of
context-sensitive formal languages that include all the context-free and regular
languages, as well as other classes such as the tree-adjoining grammars, see \cite{VSW}.
The MG grammars can also be characterized in terms of linear context free rewrite systems
(LCFRS), \cite{Micha}. However, as shown by Berwick in \cite{Berwick}, this equivalence 
between Stabler's computational minimalism and multiple context free grammars hides an
important difference in terms of ``succintness gap". Namely, computational minimalism is 
exponentially more succinct than otherwise equivalent multiple context free grammars,
(see \cite{Berwick} and \cite{Stabler}).
As shown in \cite{Berwick}, a similar gap exists between transformational generative grammar
and generalized phrase structure grammar. Another problem with thinking in terms of
formal languages is that they are designed to describe languages as strings (oriented sets) 
produced as ordered sequences of transitions in an automaton that
computes the language. This time-ordered description of languages hides its more
intrinsic and fundamental description in terms of structures (binary rooted trees without
an assigned planar embedding, in mathematical terms). Indeed the current form of
Minimalism not only to provides a more efficient encoding of the generative
process of syntax, but it also proposes a model where the core computational 
structure of syntax is entirely based on structures rather than on linear order.
The latter (equivalently, planar embeddings of trees) is superimposed to this core
computational structure, in a later externalization phase. Thus, the 
MG grammars description of Stabler's minimalism, comfortingly familiar as
it may appear to the mathematically minded, is in fact deceiving, since
formal languages only deal with properties of 
terminal strings, not with structure, while Minimalism is primarily concerned with structure
and does not need to incorporate linear order in its computational core mechanism.

\smallskip

The goal of the present paper is to elaborate on this difference. We also
want to stress the point that, while formal languages have long been
considered the mathematical theory of choice for application to
generative linguistics, it is in fact neither the only one nor the most
appropriate. As we argued in \cite{MCB}, the algebraic formalism of
Hopf algebras is a more suitable mathematical tool for theoretical linguistics,
and a useful way to move beyond the traditional thinking in terms of
formal languages. Thus, in the present paper, we will use this Hopf
algebras viewpoint to make comparisons between older forms of
minimalism like Stabler's and the current form based on free symmetric Merge,
and to show the advantages of the latter from this algebraic perspective.

\smallskip

We analyze Stabler's formulation of Minimalism
from the point of view of Hopf algebra structures, which we have shown
in \cite{MCB} to be an appropriate algebraic language for the formulation
of Merge in Minimalism. We show here that, in the older setting of Stabler's
Computational Minimalism, Internal and
External Merge correspond to very different types of mathematical
structures. External Merge is expressible in terms of the notion
of ``operated algebra", while Internal Merge can be described in terms
of right-ideal coideals in a Hopf algebra, and corresponding
quotient right-module coalgebras. 
While these are interesting mathematical structures, the very different
form of Internal and External Merge makes it difficult to reconcile them
as two forms of a single underlying basic Merge structure.
This is unsatisfactory, because linguistically one expects Internal
and External merge to be manifestations of the same
fundamental computational principle. 

\smallskip

We then show that the mathematical formulation of the New Minimalism
that we introduced in \cite{MCB} completely bypasses this problem,
by directly presenting a unified framework for both Internal and External Merge.

\smallskip

We also further discuss our proposed model for
externalization of \cite{MCB}, by comparing it with proposed 
alternative models based on possible constructions of planarizations
of trees, independent of syntactic parameters. We outline some
difficulties with such constructions, at the level of the underlying
algebraic structure. 

\medskip
\section{Stabler's Minimalism and the Loday--Ronco Hopf algebra}

In this section we first recall some mathematical structures, in particular the
Loday--Ronco Hopf algebra of {\em planar} binary rooted trees, and the explicit
form of product and coproduct. In \S \ref{HopfDefSec} we recall the basic
definition and properties of Hopf algebras and in \S \ref{BinRootTreesSec}
we introduce the specific case of binary planar rooted trees, with the
product and coproduct described in \S \ref{ProCoprodSec}. 

\smallskip

It is important to point out here that one way in which older formulation
of Minimalism differ from the more recent formulation is in the fact of
considering {\em planar} trees, namely rooted trees endowed with a
particular choice of a {\em planar embedding}. Fixing the embedding
is equivalent to fixing an ordering of the leaves of the tree, which
corresponds linguistically to considering a linear order on sentences.
While this is assumed in older versions of Minimalism, the newer
version eliminates the assignment of planar structures at the level
of the fundamental computational mechanism of Merge, relegating the
linear ordering to a later externalization procedure that interfaces
the fundamental computational mechanism of Merge (where ordering
is not assumed) with the sensory-motor system, where ordering 
is imposed in the externalization of language into speech, sign,
writing. The difference between imposing a priori a choice of
planar embeddings on trees or not leads to significantly different
mathematical formulations of the resulting Hopf algebras and
Merge operations. 

\smallskip

Another very significant difference between older versions
of Minimalism and the newer formulation is the calculus of
labels of syntactic features associated to trees, and how
such labels determine and are determined by the 
transformation implemented by Internal and External Merge.
We introduce labeling in \S \ref{LabelSec}. For a
discussion of {\em labels and projection} in the different 
versions of Minimalism see \cite{Chomsky13}. 

\smallskip

We show in \S \ref{OldIntExtSec} how in older formulations
of Minimalism such as Stabler's, the conditions on planarity
and labels impose conditions on the applicability of Merge
operations, for both External and Internal Merge, and that
this makes the operations only partially defined on specific
domains, reflecting conditions on matching/related labels 
and on projection. We discuss the specific case of
External Merge in \S \ref{OldExtMergeSec} and of
Internal Merge in \S \ref{OldIntMergeSec}. 

\smallskip

In \S \ref{IntMergeCoalgSec} and \S \ref{IntMergeAlgSec} we describe the mathematical
structure underlying the formulation of Internal Merge in the
older versions of Minimalism. In \S \ref{IntMergeCoalgSec} we show that 
the issue of labels and domains requires a modification of the 
Loday--Ronco coproduct, while in \S \ref{IntMergeAlgSec} we show that
a similar modification is required for the product, and that the problem of
heads and projection results in the fact that the domain of Internal Merge is
a right ideal but not a left ideal with respect to the product, which is a
coalgebra with resoect to the modified coproduct. 

\smallskip

Moreover, in \S \ref{IterateIntSec} we show that the presence of
domains that make Merge partially defined, due to labeling conditions,
creates an additional mathematical complication related to the
iterated application of Merge, as composition of partially defined
operations. 

\smallskip

We observe in \S \ref{CoidealSymmSec} that a significant difference
with respect to the Hopf algebra formulation of the new Minimalism
we described in \cite{MCB} arises in comparison with the analogous
structure in fundamental physics. In the case of the old Minimalism
discussed here the Internal Merge has an intrinsically asymmetric
structure, because of constraints coming from labeling and projection
and from the imposition of working with trees with planar structures.
Because of this, instead of finding Hopf ideals as in the setting
of theoretical physics, one can only obtain right-ideal coideals,
which correspond to a much weaker form of ``generalized quotients"
in Hopf algebra theory. 

\smallskip

We then discuss in more detail the mathematical structure of the
old formulation of External Merge in \S \ref{PartOpSec} and
\S \ref{ExtPartOpSec}. We show that these exhibit a
very different mathematical structure with respect to Internal Merge.
It can be described in terms of the notion of ``operated algebra", where
again one needs to extend the notion to a partially defined version
because of the conditions on domains coming from the labels and
projection problem.

\subsection{Preliminaries on Hopf algebras of rooted trees}\label{HopfDefSec}

We begin by reviewing the main constructions of Hopf algebras of rooted trees, focusing on
the {\em Loday--Ronco Hopf algebra}, which is based on 
{\em planar binary rooted trees}. 

\smallskip

Associative algebras can be regarded as the natural algebraic structure that describes linear
strings of letters in a given alphabet with the operation of concatenation of words. On the
other hand, in Linguistics the emphasis is put preferably on the syntactic trees that provide
the parsing of sentences, rather than on the strings of words produced by a given grammar.
When dealing with trees instead of strings as the main objects, the appropriate algebraic
formalism is no longer associative algebras, but Hopf algebras and more generally operads. 
These are mathematical structures essentially designed for the parsing of hierarchical compositional
rules, by encoding (in a coproduct operation) all the possible ``correct parsings", that is, all 
available decompositions of an element into its possible building blocks. 

\smallskip

We recall here some general facts about Hopf algebras that we will be using in the following. 

\smallskip

A Hopf algebra $\cH$ is a vector space over a field $\K$, endowed with 
\begin{itemize}
\item a multiplication $m: \cH \otimes_\K \cH \to \cH$;
\item a unit $u: \K \to \cH$;
\item a comultiplication $\Delta: \cH \to \cH \otimes_\K \cH$;
\item a counit $\epsilon: \cH \to \K$;
\item an antipode $S : \cH \to \cH$
\end{itemize}
which satisfy the following properties. 
The multiplication operation is associative, namely the
following diagram commutes
$$ \xymatrix{ \cH\otimes_\K \cH\otimes_\K \cH \ar[r]^{m\otimes id} \ar[d]^{id\otimes m} & \cH \otimes_\K \cH \ar[d]^m \\
\cH \otimes_\K \cH \ar[r]^m & \cH } $$
and the unit and multiplication are related by the commutative diagram
$$ \xymatrix{ & \cH \otimes_\K \cH  \ar[dd]^m & \\
\K \otimes_\K \cH \ar[ur]^{u\otimes id} \ar[dr] & & \cH \otimes_\K \K \ar[ul]^{id\otimes u} \ar[dl] \\
& \cH & } $$
where the two unmarked downward arrows are the identifications induced by scalar multiplication. 
The coproduct is coassociative, namely the following diagram commutes
$$ \xymatrix{ \cH \otimes_\K \cH \otimes_\K \cH & \cH \otimes_\K \cH \ar[l]^{\Delta \otimes id} \\
 \cH \otimes_\K \cH \ar[u]^{id\otimes \Delta} & \cH \ar[l]^\Delta \ar[u]^\Delta } $$
 and the comultiplication and the counit are related by the commutative diagram
$$ \xymatrix{ & \cH \otimes_\K \cH \ar[dl]^{\epsilon \otimes id} \ar[dr]^{id\otimes \epsilon} \\
\K \otimes_\K \cH & & \cH \otimes_\K \K \\
& \cH \ar[ul] \ar[uu]^\Delta \ar[ur] & } $$ 
where the two unmarked upward arrows are the identifications given, respectively, by
$x\mapsto 1_\K \otimes x$ and $x\mapsto x \otimes 1_\K$, with $1_\K$ the unit of the field $\K$.  
Moreover, $\Delta$ and $\epsilon$ are algebra homomorphisms, and $m$ and $u$ are coalgebra
homomorphisms. The antipode $S : \cH \to \cH$ is a linear map such that the following diagram
also commutes
$$ \xymatrix{ \cH \otimes_\K \cH \ar[r]^m & \cH & \cH \otimes_\K \cH \ar[l]^m \\
\cH \otimes_\K \cH \ar[u]^{id\otimes S} & \cH \ar[l]^\Delta \ar[u]^{u\circ \epsilon} \ar[r]^\Delta & \cH \otimes_\K \cH 
\ar[u]^{S\otimes id}
} $$ 

\smallskip

In any graded bialgebra it is possible to construct an antipode map inductively by
\begin{equation}\label{antipode}
 S(X) = -X - \sum S(X') X'', 
 \end{equation}
for any element $X$ of the bialgebra, with coproduct 
$\Delta(X)= X \otimes 1 + 1 \otimes X + \sum X' \otimes X''$,
where the $X'$ and $X''$ are terms of lower degree. Thus, in the following
we will focus only on the bialgebra structure and not discuss the
antipode map explicitly.

\smallskip

Heuristically, what these properties of a Hopf algebra describe can be summarized as follows.
As we will see in the specific example of trees below, as a vector space $\cH$ consists of formal
linear combinations of a specific class of objects (planar binary rooted trees, Feynman graphs,
etc.). The operations will be defined on these generators and extended by linearity. 
The multiplication structure is a combination operation and the coproduct structure is a decomposition
operation that lists all the possible different decompositions (parsings).
The antipode is like a group inverse and it establishes the compatibility between
multiplication and comultiplication, unit and counit. 

\medskip
\subsection{Binary rooted trees}\label{BinRootTreesSec}

A rooted tree $T$ is a finite graph whose geometric realization is simply connected (no loops),
defined by a set of vertices $V$, with a distinguished element $v_r\in V$, the root vertex, 
and a set of edges $E$, which we can assume oriented with the uniquely defined orientation
away from the root. The source and target maps $s,t: E \to V$ assign to each edge of the
tree its source and target vertices. The leaves of the tree are the univalent vertices. They
are also the sinks (no outgoing edges). A rooted tree is planar if it is endowed with an embedding
of its geometric realization in the plane. Assigning planar embedding is equivalent to assigning 
a linear ordering of the leaves. 
The tree is vertex-decorated if there is a map $L_V: V \to D_V$ to a (finite) set. It is edge-decorated
if similarly there is a map $L_E: E \to D_E$ to a finite set of possible edge decorations. We
will assume the trees are vertex-decorated, and we will simply refer to them as decorated trees. 

\smallskip

An {\em admissible cut} $C$ on a rooted tree $T$ is an operation that 
\begin{enumerate}
\item selects a number of
edges of $T$ with the property that every oriented path in $T$ from the root to one of the
leaves contains at most one of the selected edges
\item removes the selected edges.
\end{enumerate}
The result of an admissible cut is a disjoint union of a tree $\rho_C(T)$ that contains
the root vertex and a forest $\pi_C(T)$, that is, a disjoint union $\pi_C(T)=\cup_i T_i$ 
of planar trees, where each $T_i$ has a unique source vertex (no incoming edges), which
we select as root vertex of $T_i$,
\begin{equation}\label{CutT}
C(T) = \rho_C(T)\cup \pi_C(T). 
\end{equation}
An elementary admissible cut (or simply {\em elementary cut}) is a cut $C$ consisting of a single edge. 

\medskip
\subsection{The Loday--Ronco Hopf algebra of binary rooted trees}\label{LodayRoncoSec}

We note right away that this algebraic structure is
different from that we considered in \cite{MCB}. Since older
forms of Minimalism, such as Stabler's version that we discuss here,
incorporate linear ordering, it is necessary to work with binary rooted trees
with an assigned choice of planar embedding (linear ordering of the leaves). 
This has immediate consequences on the type of algebraic structure that 
we need to work with, as the possible forms of product and coproduct 
operations are affected by the presence of this linear ordering. Another
difference with respect to our setting in \cite{MCB} is that in the new
Minimalism, Merge acts on workspaces (a Hopf algebra of binary forests
with no planar embeddings), while in the old Stabler Minimalism, Merge
acts on (a Hopf algebra of) planar binary rooted trees.

\smallskip

Consider the vector space $\cV_k$ 
spanned by the planar binary rooted trees $T$ with $k$ internal vertices
(equivalently, with $k+1$ leaves). It has dimension 
$$ \dim \cV_k = (\# D_V)^k \frac{(2k)!}{k! (k+1)!}, $$
where $\# D_V$ is the cardinality of the set $D_V$ of possible vertex labels.
Let $\cV =\oplus_{k\geq 0} \cV_k$, with $\cV_0=\Q$.
For a given label $d\in D_V$ the grafting operator $\wedge_d$ is defined as
\begin{equation}\label{bingraft}
 \wedge_d : \cV \otimes \cV \to \cV,  \ \ \  T_1\otimes T_2 \mapsto T= T_1 \wedge_d T_2   \, , 
\end{equation} 
with $\wedge_d: \cV_k \otimes \cV_\ell \to \cV_{k+\ell-1}$,
that attaches the two roots $v_{r_1}$ of $T_1$ and $v_{r_2}$ of $T_2$ to a single root vertex 
$v$ labelled by $d\in D_V$. 

\smallskip

One also introduces the following associative concatenation operations on planar binary rooted trees: 
given $S$ and $T$, the tree $S\backslash T$ ($S$ under $T$) is obtained by grafting the 
root of $T$ to the rightmost leaf of $S$, while $T/S$ ($S$ over $T$) is the tree obtained 
by grafting the root of $T$ to the leftmost leaf of $S$.  The grafting operation \eqref{bingraft}
is related to these concatenations by
\begin{equation}\label{wedgeT12}
T_1 \wedge_d T_2   = T_1 / S \backslash T_2, 
\end{equation}
where $S$ is the planar binary tree with a single vertex decorated by $d\in D_V$.
Each planar binary rooted tree is described as a grafting $T=T_\ell \wedge_d T_r$,
of the trees stemming to the left and right of the root vertex. We remove the
explicit mention of the vertex decorations when not needed.

\smallskip

The Loday--Ronco Hopf algebra $\cH_{LR}$ of planar binary rooted trees is
obtained from the vector space $\cV$ by defining a multiplication and a
comultiplication inductively by degrees. 
For trees $T=T_\ell \wedge T_r$ and $T'=T'_\ell \wedge T'_r$ the product defined in
\cite{LoRo} can be built inductively using the property
$$ T\star T' = T_\ell \wedge (T_r \star T') + (T\star T'_\ell) \wedge T'_r , $$
with the tree consisting of a single root vertex $\bullet$ as the unit. The coproduct
similarly can be built from lower degree terms by the property
$$ \Delta(T) = \sum_{j,k} (T_{\ell,j} \star T_{r,k}) \otimes (T'_{\ell,n-j} \wedge T'_{r,m-k}) + T \otimes \bullet $$
where $T=T_\ell \wedge T_r$ and $\Delta(T_\ell)=\sum_j T_{\ell,j}\otimes T'_{\ell,n-j}$ and
$\Delta(T_r)=\sum_k T_{r,k} \otimes T'_{r, m-k}$, for $T_\ell \in \cV_n$ and $T_r\in \cV_m$.

\smallskip

In \cite{LoRo} this Hopf algebra $\cH_{LR}$ of planar binary rooted trees is described in terms of
the Hopf algebra structure on the group algebra $\Q[S_\infty]=\oplus_n \Q[S_n]$ 
of the symmetric group. Namely, the inclusion of the Hopf algebra of non-commutative 
symmetric functions in the Malvenuto-Reutenauer Hopf algebra of permutations 
factors through the Loday-Ronco Hopf algebra of planar binary rooted trees, see
also \cite{AguSot}. In \cite{BroFra} a version of the Loday-Ronco Hopf
algebra of planar binary rooted trees was used for the renormalization of 
massless quantum electrodynamics and and explicit isomorphisms between the Loday-Ronco
Hopf algebra of planar binary rooted trees and the (noncommutative) 
Connes--Kreimer Hopf algebra of renormalization were constructed 
in \cite{AguSot}, \cite{Holt1}, \cite{Foissy2}.

\medskip
\subsection{Graphical form of coproduct and product}\label{ProCoprodSec}

It is shown in \cite{AguSot} that the coproduct and product of the 
Loday-Ronco Hopf algebra of planar binary rooted trees can be
conveniently visualized in the following way.

\smallskip

Given a tree $T$, one subdivides it into two parts by cutting along the path from
one of the leaves to the root (illustration from \cite{AguSot})
\begin{center}
\includegraphics[scale=0.5]{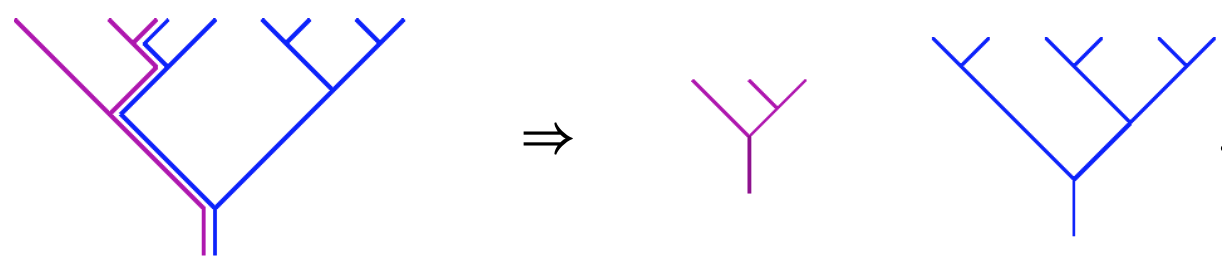} 
\end{center}
and the coproduct then takes the form of a sum over all such decompositions
\begin{equation}\label{coprodLR}
\Delta(T)=\sum T'\otimes T''
\end{equation}
where $T',T''$ are, respectively the parts of $T$ to the left and right of the 
path from a leaf to the root and all choices of leaves are summed over. 

\smallskip

In order to form the product $T_1 \star T_2$, suppose that $T_1$ has $n_1+1$ leaves
and $T_2$ has $n_2+1$ leaves. Consider again subdivisions of the tree $T_1$ obtained
by cutting along paths from the leaves to the root, including trees consisting of just the
path itself, with the cuts chosen so as to obtain $n_2+1$ subtrees of $T_1$ 
(illustration from \cite{AguSot} in a case with $n_1=5$, $n_2=3$)
\begin{center}
\includegraphics[scale=0.5]{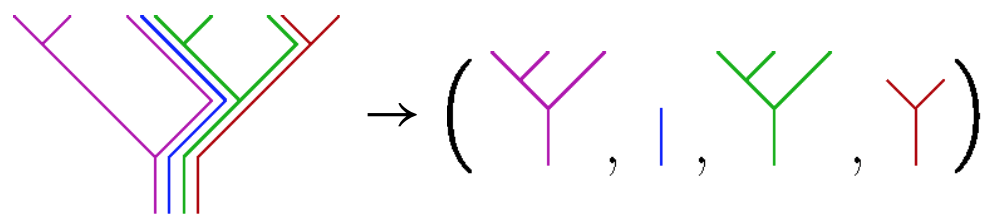}  \\
\includegraphics[scale=0.5]{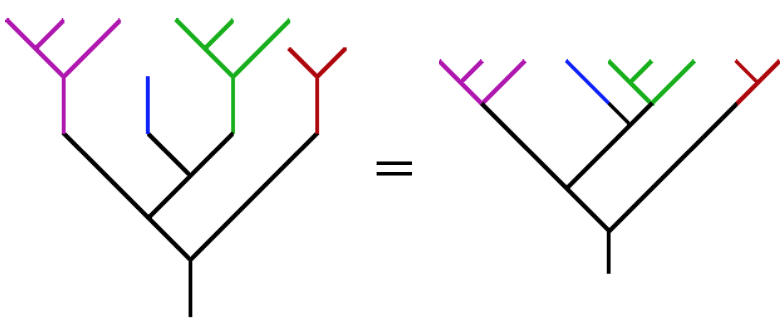} 
\end{center}

\smallskip

We write $\cV^{(k)}$ for the $\Q$-vector space spanned by the planar binary rooted trees with $k$ leaves,
and we write the usual operadic composition operation of grafting roots to leaves as above
in the form
\begin{equation}\label{operad}
\gamma: \cV^{(k_0)}\times \cdots \times \cV^{(k_n)} \times \cV^{(n+1)} \to \cV^{(k_0+\cdots + k_n)}
\end{equation}
where $\gamma(T_0,\ldots, T_n; T)$ is the tree obtained by
grafting the trees $T_i\in \cV^{(k_i)}$ in collection $(T_0,\ldots, T_n)$  to the tree $T\in \cV^{(n+1)}$, 
by attaching the root of $T_i$ to the $i$-th leaf of $T$, as illustrated above.

\smallskip

The product in $\cH_{LR}$ is then written in the form
\begin{equation}\label{prodLR}
T \star T' =\sum_{(T_0,\ldots, T_n)} \gamma(T_0,\ldots, T_n; T')
\end{equation} 
with $n+1$ the number of leaves of $T'$, and the sum over subdivisions into subtrees 
extracted according to the rule described above. 

\medskip
\subsection{Labelled trees}\label{LabelSec}

The previous subsections only dealt with general mathematical
formalism about planar binary rooted trees. We now consider
more specifically the case of Stabler's Computational Minimalism.

\smallskip

In Stabler's formulation, one considers planar binary rooted trees such as
the following example (from \cite{Stabler}).
\begin{center}
\includegraphics[scale=0.5]{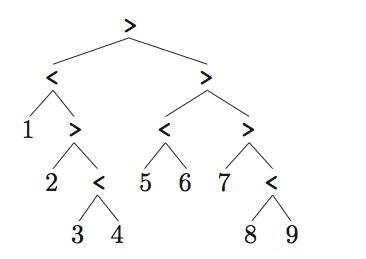} 
\end{center}
where the ordering of the leaves determines and is determined by
the planar embedding of the tree, and the labels at the inner vertices
serve the purpose of pointing toward the head. 

\smallskip

More precisely, all internal vertices and the root
vertex are labelled by symbols in the set $\{ >, < \}$. The purpose
of the labels $>$ and $<$ is to identify where the {\em head} of the tree is: 
in the example above the head is the leaf vertex number $8$. 
In addition to these labels, one also considers a finite set of syntactic features $X\in \{ N,V,A,P,C,T,D,\ldots \}$
and ``selector" features denoted by the symbol $\sigma X$ for a head selecting a phrase $XP$. More
generally we can have labels that are strings (ordered finite sets) $\alpha=X_0 X_1 \cdots X_r$ of syntactic
features as above. We also consider labels given by letters $\omega$ and $\bar\omega$ 
that stand for ``licensor" and ``licensee".
(Note: in \cite{Stabler} the notation $= X$ is used for feature selector instead of our $\sigma X$,
and the notation $\pm X$ is used for licensor/licensee pairs,
but we prefer to avoid using mathematical notation with a meaning different from its generally accepted one,
to avoid confusion, hence we will use the letters $\sigma$ and $\omega,\bar\omega$ instead.)

\smallskip

We write the result of external merge
applied to constituents $\alpha$ and $\beta$ as in \cite{Stabler}, as the labelled planar binary rooted tree
\begin{center}
\Tree [ .$<$  $\alpha$ $\beta$ ]  \ \ \ \  \ \ \ \   \ \ \ \   rather than  \Tree [ .$\alpha$ $\alpha$ $\beta$ ]
\end{center}
This means that we consider as generators of the Loday--Ronco Hopf algebras the binary trees
with labels that include the syntactic features, as well as symbols $<$ and $>$, as in \cite{Stabler}, 
used to point towards the head when merge is applied. 

\smallskip

By \eqref{wedgeT12} it is clear that the head of the tree $T_1 \wedge_{>} T_2$
is the head of $T_2$ and the head of the tree $T_1 \wedge_{<} T_2$ is the
head of $T_1$. 

\smallskip

Recall also that a {\em maximal projection} in $T$ is a subtree of $T$ that is not a proper
subtree of any larger subtree with the same head. As pointed out in \cite{Stabler},
in the example illustrated above, the leaves 
$\{2,3,4\}$ determine a subtree with head vertex the leaf numbered $3$, but any larger
subtree in $T$ would have a different head, hence this subtree is a maximal projection.
So is the subtree determined by the leaves $\{ 5,6 \}$, for instance. 

\smallskip

We will write $\cH_{ling}$ for the Loday-Ronco Hopf algebra $\cH_{LR}$ with the
choice of labels described above. We will write $\cV_{ling}$ for the underlying $\Q$-vector
space spanned by the binary trees with the labeling described above.

\smallskip

As we pointed out already, the choice of working with {\em planar} binary rooted 
trees corresponds to an assigned linear ordering of its leaves,
which in turn is the linear ordering of the resulting sentence when spoken or read.
Thus, by working with planar trees we are assuming that linear ordering is determined for the
result of Merge. This is the same assumption made in \cite{Stabler}. This is the crucial
point in the comparison with the new Minimalism, and we will discuss it more
explicitly in Section~\ref{NewMinSec} below.

\medskip
\subsection{Old formulation of External and Internal Merge}\label{OldIntExtSec}

We now discuss the external and internal merge operations in the old
formulations of minimalism. We first describe
the operations at the level of the underlying combinatorial tree, and then we
give the more precise constraints on when the operations can be applied, based
on the labels of the trees involved.

\smallskip

Thus, the way we should view external and internal merge is as {\em partially defined
operations} 
$$ \cE: \cV_{ling} \otimes \cV_{ling} \to \cV_{ling} $$
$$ \cI : \cV_{ling} \to \cV_{ling} $$
where we assume that the operations defined on generators are
extended by (bi)linearity, so that they are defined on linear subspaces 
$$ {\rm Dom}(\cE) \subset \cV_{ling} \otimes \cV_{ling}  $$
$$ {\rm Dom}(\cI) \subset \cV_{ling} $$
that we will describe more precisely below. 

\smallskip  

At the level of the underlying combinatorial trees external and internal merge are
simply defined as the following operations.

\smallskip

The combinatorial structure of the external merge is given by
\begin{equation}\label{extmerge}
\cE(T_1 \otimes T_2)=\left\{ \begin{array}{ll} \bullet \wedge T_2 & T_1=\bullet \\[3mm]
T_2 \wedge T_1 & \text{otherwise,}
\end{array}\right.
\end{equation}
where $\bullet$ denotes the tree consisting of a single vertex, and the $\wedge$
operation is the grafting operation $\wedge_d$ defined as in \eqref{bingraft},
where the specific label $d \in \{ >, < \}$ according to a rule that will be discussed 
more in detail in \S \ref{OldExtMergeSec} below.

\smallskip

The combinatorial structure of the internal merge is given by
\begin{equation}\label{intmerge}
\cI(T)= \pi_C(T) \wedge \rho_C(T),
\end{equation}
where $C$ is an elementary admissible cut of $T$ with $\rho_C(T)$ the
remaining pruned tree that contains the root of $T$ and $\pi_C(T)$ the
part that is severed by the cut (which in the case of an elementary cut
is itself a tree rather than a forest). Again $\wedge$ is as in \eqref{bingraft},
with a label that will be made more precise below, and the choice of the
elementary cut will also depend on conditions on tree labels, see \S \ref{OldIntMergeSec}. 

\smallskip

We now look more precisely at the structure of external and internal merge
and at their domains of definition. It should be noted how working with
planar structures and with conditions of matching/related labels makes
the underlying mathematical operations more difficult to describe, as
they are only defined on specific domains and with additional conditions.

\medskip
\subsection{Old form of External Merge}\label{OldExtMergeSec}

The more precise definition of the External Merge in the older forms of Miimalism 
is given as follows (where we are
adapting the description of \cite{Stabler} to our terminology and notation).

\smallskip

As in \cite{Stabler}, we use the notation $T[\alpha]$ for a tree where the head is
labelled by an ordered set of syntactic features starting with $\alpha$. 
 The label $\alpha$ 
consists of a string of syntactic features of the form
$$  \alpha= X_0 X_1\cdots X_r \ \ \ \text{ or } \ \ \ \alpha= \sigma X_0 X_1\cdots X_r, $$
with $\sigma$ the selector symbol. 

\smallskip

We introduce the notation $\hat \alpha$ for the string obtained from $\alpha$ after 
removing the first feature (other than the selection symbol). Namely, for $\alpha=X_0 X_1\cdots X_r$
or $\alpha=\sigma X_0 X_1\cdots X_r$, we have $\hat \alpha=X_1\cdots X_r$.

\smallskip

The external
merge $T=\cE(T_1[\sigma \alpha], T_2[\alpha])$ of two trees $T_1[\sigma \alpha]$ and $T_2[\alpha]$ is given by 
\begin{equation}\label{extmergefull}
 \cE(T_1[\sigma \alpha], T_2[\alpha]) = \left\{ \begin{array}{ll} T_1[\widehat{\sigma \alpha}] \wedge_< T_2[\hat \alpha]  & |T_1|=1 \\
T_2[\hat \alpha] \wedge_>  T_1[\widehat{\sigma \alpha}]  & |T_1|>1. \end{array} \right. 
\end{equation}

\smallskip

This clearly agrees with \eqref{extmerge} at the level of the underlying combinatorial trees. 
The main difference with \eqref{extmerge} is that here the operation can be performed only
if the heads are decorated with syntactic features $\sigma \alpha$ and $\alpha$, respectively. 
Thus the domain of definition ${\rm Dom}(\cE)\subset \cV_{ling}\otimes \cV_{ling}$ 
of the external merge operation is the vector space on the set of generators 
\begin{equation}\label{DomExtMerge}
{\rm Dom}(\cE)={\rm span}_\Q \{ (T_1[\beta],T_2[\alpha]) \,|\, \beta=\sigma \alpha  \}.
\end{equation}

\medskip
\subsection{Old version of Internal Merge} \label{OldIntMergeSec} 

In the older formulations of Minimalism, one considers a 
second Merge operation, considered as distinct from External Merge, namely the {\em Internal Merge}.
We again adapt to our notation and terminology the formulation given in  \cite{Stabler}. We use
the same notation as above for $T[\alpha]$ with $\alpha=X_0X_1\ldots X_r$ or 
$\alpha=\sigma X_0X_1\ldots X_r$ a string of syntactic features possibly starting with a selector,
and we write $\hat\alpha=X_1\ldots X_r$. The internal merge 
is again modeled on the basic grafting operation of planar rooted binary trees $\wedge_d$,
but this time its input is a single tree $T[\alpha]$.

\smallskip

Given a planar binary rooted tree $T$ containing a subtree $T_1$, and given another 
binary rooted tree $T_2$, we denote as in \cite{Stabler} by $T\{ T_1 \to T_2\}$ the
planar binary rooted tree obtained by removing the subtree $T_1$ from $T$ and replacing it
with $T_2$. In particular, we write $T\{ T_1 \to \emptyset \}$ for the tree obtained by
removing the subtree $T_1$ from $T$. Note that, in order to ensure that the result of
this operation is still a tree, we need to assume that if an internal vertex of $T$ belongs
to the subtree $T_1$, then all oriented paths in $T$ from this vertex to a leaf also belong
to $T_1$, where paths in the tree are oriented from root to leaves. We only consider subtrees
that have this property. When it is necessary to stress this fact, we will refer to such subtrees
as {\em complete subtrees}.
Note that these are exactly the subtrees that can be obtained by
applying a single elementary admissible cut $C$ 
to the tree $T$. The tree $T\{ T_1 \to \emptyset \}$ is then the same as the tree 
$\rho_C(T)$. 

\smallskip

Consider a tree  $T[\alpha]$ where $\alpha= X_0\cdots X_r$ or $\alpha=\sigma X_0\cdots X_r$ or $\alpha=\omega X_0\cdots X_r$ or $\alpha=\bar \omega  X_0\cdots X_r$. 

\smallskip

The domain ${\rm Dom}(\cI)\subset \cV_{ling}$ of the internal merge is given by the subspace
\begin{equation}\label{DomIntMerge}
{\rm Dom}(\cI) ={\rm span}_\Q\{ T[\alpha] \,|\, \exists T_1[\beta] \subset T[\alpha], \text {with } \beta=\bar\omega X_0 \hat \beta, \, \alpha=\omega X_0 \hat \alpha \}\, , 
\end{equation}
where $T_1\subset T$ is a subtree (in the sense specified above), and
\begin{equation}\label{IntMerge}
\cI(T[\alpha])= T_1^M[\hat \beta] \wedge_> T\{ T_1[\beta]^M \to \emptyset \} = \pi_C(T) \wedge_> \rho_C(T),
\end{equation}
where $C$ is the elementary admissible cut specified by the subtree $T_1^M$ (the maximal projection
of the head of $T_1$) and the condition (given by the matching labels $\omega X_0$ and $\bar\omega X_0$)
that $T[\alpha]\in {\rm Dom}(\cI)$.  The head of $\pi_C(T) \wedge_> \rho_C(T)$ gets the label $\hat\alpha$.

\smallskip

Stabler's External and Internal Merge, that we recalled here in \eqref{extmergefull}
and \eqref{IntMerge}, have issues at the linguistics level, such as problems with 
potentially unlabelable exocentric constructions.  For example (as observed by
Riny Huijbregts)
Internal Merge as in \eqref{IntMerge} gives $\{XP, YP \}$ results, and labeling cannot
be assigned, unless a further application of Internal Merge takes $XP$ or $YP$ in a 
criterial position, where structural agreement, SPEC-Head agreement, can take place 
between specifier and head. In \cite{Stabler}, two further variants of 
External and Internal Merge are introduced in order to deal with ``persistent features".
This results in what is referred to in \cite{Stabler} as ``conflated minimalist grammars" (CMGs).
These additional forms of External and Internal Merge, however, do not solve the
problem mentioned above and further complicate the structure, so we will focus here
on analyzing the algebraic structure determined by just the forms
\eqref{extmergefull} and \eqref{IntMerge} of Stabler's External and Internal Merge.

\medskip
\subsection{Old version of Internal Merge and the coalgebra structure}\label{IntMergeCoalgSec}

As we have discussed above, the internal merge is only partially defined on $\cV_{ling}$
because it requires the existence of matching conditions on the label, which determine
the domain ${\rm Dom}(\cI)\subset \cV_{ling}$. We now discuss now how the domain 
${\rm Dom}(\cI)$ and the internal merge operation $\cI$ behave with respect to the
coproduct of the Hopf algebra $\cH_{ling}$.

\smallskip

Consider the coproduct $\Delta(T)$ of a tree $T\in {\rm Dom}(\cI)$. This is given by
the sum of decompositions $\Delta(T)=\sum T' \otimes T''$ as illustrated in \eqref{coprodLR}. 
In each of these decompositions one side will contain the head of the tree $T$ 
(or possibly both sides if the leaf used to cut the tree happens to be also the head). 

If both the head of $T$ and the head of $\pi_C(T)$ are contained in the same
side of the partition, then that side still belongs to ${\rm Dom}(\cI)$.
Thus, these pieces of the coproduct are in
${\rm Dom}(\cI)\otimes \cH_{ling} + \cH_{ling} \otimes {\rm Dom}(\cI)$. The remaining
terms, with the heads of $T$ of $\pi_C(T)$ on different sides of the decomposition 
only belong in general to $\cH_{ling} \otimes \cH_{ling}$ and fall outside of the domain of the internal merge.

\smallskip

This suggests a modification of the coproduct on $\cH_{ling}$ with respect to the
usual Loday-Ronco coproduct \eqref{coprodLR}. For a generator $T$ that is
not in ${\rm Dom}(\cI)$ we just set $\pi_C(T)=T$, that is, no cut is performed. 

\smallskip

For $T\in {\rm Dom}(\cI)$, let $h(T)$ and $h(\pi_C(T))$ be the leaves of $T$ that are 
the head of $T$ and the head of $\pi_C(T)$, respectively.
If $T$ is in ${\rm Dom}(\cI)$ then $C$ is, as above, the elementary admissible
cut determined by the label conditions. Let $\cP_\cI(T)$ denote the set of bipartitions
\begin{equation}\label{bipartheads}
\cP_\cI(T)=\{ T=(T',T'') \,|\, ( h(T)\in T' \text{ and } h(\pi_C(T)) \in T') \text{ or } (h(T)\in T'' \text{ and } h(\pi_C(T)) \in T'') \}\, . 
\end{equation}
For trees outside the domain ${\rm Dom}(\cI)$ the set $\cP_\cI(T)$ just counts all bipartitions
as we have set $\pi_C(T)=T$. 

\smallskip

We then define the modified coproduct
\begin{equation}\label{coprodDom}
\Delta_\cI(T):=\sum_{(T',T'')\in \cP_\cI(T)} T'\otimes T'',
\end{equation}
which is the same as \eqref{coprodLR} outside of ${\rm Dom}(\cI)$.

\smallskip

With this modified coproduct ${\rm Dom}(\cI)$ is a {\em coideal} of the coalgebra $\cH_{ling}$, namely 
\begin{equation}\label{Hopfideal}
\Delta_\cI ({\rm Dom}(\cI)) \subset {\rm Dom}(\cI)\otimes \cH_{ling} + \cH_{ling} \otimes {\rm Dom}(\cI) \, . 
\end{equation}

\medskip
\subsection{Old version of Internal Merge and the algebra structure}\label{IntMergeAlgSec}

We now discuss the behavior of internal merge with respect to the product structure of $\cH_{ling}$.
In this case, arguing in a similar way as for the coproduct above, there is a modification $\star_\cI$ of the
original product of $(\cH_{LR}, \star)$ that remains the same outside of ${\rm Dom}(\cI)$ and takes into
account the additional data in ${\rm Dom}(\cI)$ and that has the effect of making ${\rm Dom}(\cI)$
into a {\em right ideal} with respect to the algebra $(\cH_{ling}, \star_\cI)$. However, ${\rm Dom}(\cI)$
is not a left ideal. 

\smallskip

We define the modified product $\star_\cI$ on $\cH_{ling}$ in the following way.
Let $h(T)$ denote the head of $T$. Given trees $T,T'$ where $T'$ has $n+1$ leaves,
consider the set $\cP_\cI(T,T')$ given by decompositions $(T_0,\ldots, T_n)$ of $T$
as above with
\begin{equation}\label{partsTT}
 \cP_\cI(T,T'):=\{ (T_0,\ldots, T_n)\,|\, h(T) \text{ and } h(\pi_C(T)) \in T_{h(T')} \} \,
\end{equation}
Namely those decompositions $(T_0,\ldots, T_n)$ with the property that the head of $T$
(and the head of $\pi_C(T)$ in the case where $T\in {\rm Dom}(\cI)$) lies in the component
$T_{h(T')}$ that is grafted to the head of $T'$. In the case of $T\notin {\rm Dom}(\cI)$
the condition reduces to just $h(T)\in T_{h(T')}$. Note that it is always possible to have such
decompositions, as some of the components $T_i$ can always be taken to be copies of just
the path from a leaf to the root, so that the relevant piece of the decomposition can be placed
at the $h(T')$-position.

\smallskip

We then take the product $\star_\cI$ on $\cH_{ling}$ of the form
\begin{equation}\label{prodI}
T \star_\cI T' =\sum_{(T_0,\ldots, T_n) \in \cP_\cI(T,T')} \gamma(T_0,\ldots, T_n ; T') \,.
\end{equation}
The head of each $\gamma(T_0,\ldots, T_n ; T')$ is then the same as the head of $T$,
which we write in shorthand as $h(T\star_\cI T')=h(T)$. Moreover, by construction the
component $T_{h(T')}$ is in ${\rm Dom}(\cI)$ when $T\in {\rm Dom}(\cI)$, so that
$T\star_\cI T'$ is itself in ${\rm Dom}(\cI)$. This shows that ${\rm Dom}(\cI)$ is a right
ideal with respect to the algebra $(\cH_{ling}, \star_\cI)$,
$$  {\rm Dom}(\cI)\, \star_\cI \, \cH_{ling} \subset {\rm Dom}(\cI)\, . $$
It is, however, not a left ideal. 

\smallskip

For $T\in {\rm Dom}(\cI)$ we then have
\begin{equation}\label{intmergeprod}
 \cI(T\star_\cI T') =\sum_{(T_0,\ldots, T_n) \in \cP_\cI(T,T')}  \pi_C(T_{h(T')})\wedge_> \gamma(T_0, \ldots, \rho_C(T_{h(T')}), \ldots, T_n; T')\, . 
\end{equation} 

\smallskip

Combining the behavior with respect to the product with the previous observation 
about the coproduct we conclude that
the internal merge $\cI$ defines a right $(\cH_{ling}, \star_\cI)$-module given by the cosets
$$ \cM_\cI := {\rm Dom}(\cI) \backslash \cH_{ling} $$
where $\cM_\cI$ is also a coalgebra with the coproduct induced by $(\cH_{ling}, \Delta_\cI)$. 

\medskip
\subsection{Iterated Internal Merge}\label{IterateIntSec}

In fact, the construction above can be extended to a nested family of right-ideal coideals, 
given by the domains of the iterations of the internal merge, $\cD_{N+1}\subset \cD_N$ with
$$ \cD_N := {\rm Dom}(\cI^N)\, . $$

\smallskip

When we consider repeated application of $N$ internal merge operations, starting from a given
tree $T[\alpha]$, where $\alpha$ is a finite list of features each of which can be either $X_i$ or $\omega X_i$
of $\bar\omega X_i$ as above, we need to assume the following conditions
\begin{enumerate}
\item There are $N$ subtrees $T_1,\ldots, T_N$ in $T$, each of which is a complete subtree,
in the sense specified above. 
\item Let $T_1^M, \ldots, T_N^M$ be the maximal projections of the subtrees, in the sense
recalled above. These are also complete subtrees of $T$.
\item The subtrees $T_i^M$ are disjoint. 
\end{enumerate}
In addition to these combinatorial conditions, we have a matching labels condition that
specifies the domain of the composite operation. Let $T[\alpha]$ be the given tree as a labelled
tree, and let $T_1[\beta^{(1)}],\ldots, T_N[\beta^{(N)}]$ be the labelled subtrees, with the conditions
listed above. The domain of the composite internal merge is then given by
\begin{equation}\label{DomIntMergeN}
{\rm Dom}(\cI^N) = \left\{ T[\alpha] \,\big|\,  \exists T_1[\beta^{(1)}],\ldots, T_N[\beta^{(N)}]\, \text{ with } \left\{ \begin{array}{l} \text{(1), (2), (3)
are satisfied}\\
\beta^{(1)}_0=\bar\omega X_0, \ldots,  \beta^{(N)}_0 = \bar \omega X_{N-1} \\
\alpha=\omega X_0 \omega X_1 \cdots \omega X_{N-1} \cdots
\end{array}\right\} \right\} .
\end{equation}

For a forest $F=T_1\cdots T_\ell$, we define a grafting operation $\wedge^\ell$ that consists
of the repeated application of the grafting $\wedge$ to the trees in $F$,
\begin{equation}\label{graftN}
\bigwedge^\ell F = T_1\wedge T_2 \wedge \cdots \wedge T_\ell .
\end{equation}

The conditions (1), (2), and (3) above ensure that the choice of the subtrees $T_1^M, \ldots, T_N^M$
corresponds to an admissible cut $C$ of the tree $T$, with the number of cut branches $\# C =N$
The planar binary rooted tree obtained by this operation is then
\begin{equation}\label{intMergeN}
\cI^{\# C}(T[X])=\bigwedge^{1+\# C} \left(\pi_C(T)[\hat \bY]\,\, \rho_C(T)[\hat X^N] \right),
\end{equation}
where we use the notation $\pi_C(T)[\hat \bY]$ for the forest $\pi_C(T)=T_N^M \cdots T_1^M$,
where the label $[\hat \bY]$ means 
$$ \pi_C(T)[\hat \bY] = T_N^M[\hat \beta^{(N)}] \cdots T_1^M[\hat \beta^{(1)}] $$
and the label $[\hat \alpha^N]$ of the tree $\rho_C(T)$ stands for what remains of the original label $X$ after
the initial terms $\omega X_0 \omega X_1 \cdots \omega X_{N-1}$ are removed.

\smallskip

Each $\cD_{N+1}\backslash\cD_N$ determines a coideal in the coalgebra 
$\cD_{N+1}\backslash \cH_{ling}$ and this gives a projective system of
right-module coalgebras  
$$ \cM_{\cI^N} := {\rm Dom}(\cI^N) \backslash \cH_{ling} \, . $$

\medskip
\subsection{Coideals, recursive structures, and symmetries}\label{CoidealSymmSec}

In the Minimalist Model of syntax, the Merge operator is the main mechanism for the
construction of recursive and hierarchical structures. So indeed it is
not surprising that the Hopf-algebraic formalism occurs naturally in this context,
as it is known to describe similar hierarchical structures (such as Feynman 
graphs) in fundamental physics.
In the applications of Hopf algebras to physics, especially to renormalization in
perturbative quantum field theories, Hopf ideals are closely related to the recursive 
implementation of symmetries (Ward identities) and the recursive construction of
solutions to equations of motion (Dyson-Schwinger equations), \cite{Foissy}, \cite{Yeats}.

\smallskip

We have discussed in our previous paper \cite{MCB} how the algebraic formulation
of the new Minimalism closely resembles the mathematical structure of 
Dyson-Schwinger equations in physics. However, the analogous mathematical
structure describing Merge in the old forms of Minimalism differs significantly from
that basic fundamental formulation we have in the case of the new Minimalism. 

\smallskip

The main difference, as shown above, lies in the intrinsic
asymmetry of the old form of the Merge operation, which only gives rise to 
a weaker structure than
Hopf ideals, namely the right-ideal coideals discussed in \S \ref{IntMergeAlgSec}. 
Thus, instead of a quotient Hopf algebra as in the case of implementation
of symmetries in quantum field theory, 
one only obtains a quotient right-module coalgebra. 

\smallskip

Such objects, quotient right-module coalgebras, sometimes referred to as
``generalized quotients" of Hopf algebras, do provide a suitable notion
of quotients in the case of noncommutative Hopf algebras, \cite{Masu}, \cite{Take}.
They are also studied in the context
of the theory of Hopf--Galois extensions, designed to provide good
geometric analogs in noncommutative geometry of principal bundles and
torsors in ordinary geometry, see for instance \cite{SchaSchn}. However,
the type of structure involved, in order to accommodate the requirement
of working with planar trees, becomes significantly more complicated than
in the case of free symmetric Merge of the new Minimalism.

\smallskip

We can think of the right-module coalgebras $\cM_{\cI^N}$ as a
family of geometric spaces (in a noncommutative sense) that implement the
recursive structures of syntax generated by the Merge operation, in a
sense similar to how quotients by Hopf ideals in physics recursively implement the
gauge symmetries or the recursively constructed solutions to the equations of motion. 
This, however, is a significantly weaker (and at the same time significantly more
complicated) algebraic structure than the one we see occurring
in the newer version of Minimalism, as shown in \cite{MCB} and discussed 
in \S \ref{NewMinSec} below.

\medskip
\subsection{Partial operated algebra}\label{PartOpSec}

We now discuss more in detail the mathematical structure of the old formulation of External
Merge, and we show how it creates an independent structure of a very different nature from
Internal Merge discussed in \S \ref{IntMergeAlgSec}. These very different mathematical
formulations of Internal and External Merge are another significant drawback of the older
Minimalism in comparison with the newer. 

\smallskip

We can consider the data $(\cH_{ling}, \star_\cI, {\rm Dom}(\cI), \cI)$ discussed above
as a generalization of the notion of {\em operated algebra} (see \cite{Guo}, \cite{Zhang})
where one considers data $(\cA,\star, F)$ of an algebra together with a linear operator $F: \cA\to \cA$.
Here we allow for the case where the linear operator $F: {\rm Dom}(F)\to \cA$ is only
defined on a smaller domain ${\rm Dom}(F)\subset \cA$ which is a right ideal of $\cA$. 
We can refer to this structure $(\cH_{ling}, \star_\cI, {\rm Dom}(\cI), \cI)$ as a {\em partial operated algebra}.

\smallskip

The setting of operated algebras was introduced by Rota \cite{Rota} as a way of formalizing various
instances of linear operators $F: \cA\to \cA$ on an algebra that satisfy polynomial constraints.
The simplest example of such polynomial constraints is the identity $F(a\star b)=F(a)\star F(b)$ that
makes $F$ an actual algebra homomorphism. Another example is the Leibniz rule constraint 
$F(a\star b)=a\star F(b) + F(a) \star b$ that makes $F$ a derivation. Other interesting polynomial
constraints are Rota--Baxter relations, $F(a)\star F(b)=F(a\star F(b)) + F(F(a)\star b) + \lambda F(a\star b)$,
which depending on the value of the parameter $\lambda$ can represent various types of operations
such as integration by parts, or extraction of the polar part of a Laurent series, and play an important
role in the Hopf algebra formulation of renormalization in physics. 

\smallskip

In the case of the internal merge we considered above, not only the linear operator
is only partially defined as $\cI: {\rm Dom}(\cI)\to \cH_{ling}$, but the identity \eqref{intmergeprod} 
that expresses the internal merge of a product is not directly of a simple poynomial form as in Rota's
operated algebra program. However, it appears to be an interesting question whether the
type of relation \eqref{intmergeprod} can be accommodated in terms of the approach to
Rota's program via {\em term rewriting systems} (in the sense of \cite{BaaNi}) developed
in \cite{GaGuo}, \cite{Guo}.

\smallskip

\medskip
\subsection{Old version of External merge and the Hopf algebra structure}\label{ExtPartOpSec}

We now focus on the binary operation $\cE$ of External Merge, 
defined on the
subdomain ${\rm Dom}(\cE) \subset \cH_{ling}\otimes \cH_{ling}$ 
described in \eqref{DomExtMerge}. The structure of external merge
is simpler than the internal merge discussed above, and closely related
to the usual Hochschild cocycle that defines the grafting operations in
the Hopf-algebraic construction of Dyson-Schwinger equations in physics.

\smallskip

The ``operated algebra" viewpoint we discussed briefly above is especially 
useful in analyzing the external merge, along the
lines of extending the notion of operated algebra from unitary to binary operations discussed
in \cite{Zhang}. 

\smallskip

We recall from \cite{Zhang} the notion of operated algebra based on binary operations.
These structure is called a $\vee_\Omega$-algebra in \cite{Zhang}. In our notation we use $\wedge$
instead of $\vee$ for the grafting of binary trees used in the merge operation, following the
convention of writing syntactic parsing trees with the root at the top and the leaves at the
bottom. So we are going to refer to this structure as $\wedge_\Omega$-algebra.

\smallskip

A $\wedge_\Omega$-algebra is an algebra $(\cA,\star)$ together with a family of binary operations
$\{ \wedge_\alpha \}_{\alpha \in \Omega}$ satisfying the identity
\begin{equation}\label{veeid}
 a\star b = a_1 \wedge_\alpha (a_2\star b) + (a\star b_1) \wedge_\alpha b_2 \, , 
\end{equation} 
for $a=a_1 \wedge_\alpha a_2$ and $b=b_1\wedge_\alpha b_2$.
A $\wedge_\Omega$-bialgebra (or Hopf algebra) $(\cH, \Delta, \star, \wedge_\Omega)$ 
is a bialgebra (or Hopf algebra) that is also a $\wedge_\Omega$-algebra.
It is a {\em cocycle} $\wedge_\Omega$-Hopf algebra if the binary operations $\wedge_\Omega$
also satisfy the cocycle identity
\begin{equation}\label{cocycle}
 \Delta (a \wedge_\alpha b) =(a \wedge_\alpha b) \otimes 1 + (\star \otimes \wedge_\alpha) \circ \tau (\Delta(a)\otimes \Delta(b)) \, ,
\end{equation} 
where $\tau: \cH\otimes\cH\otimes \cH\otimes \cH \to \cH\otimes\cH\otimes \cH\otimes \cH$ is the
permutation that exchanges the two middle factors. 

\smallskip

It is shown in  \cite{Zhang} that the Loday-Ronco Hopf algebra $\cH_{LR}$ of
planar binary rooted trees with the binary operations $\wedge_\alpha$ that
graft two trees $T_1, T_2$ as the left and right subtrees at a new binary root
with label $\alpha$ is a cocyle $\wedge_\Omega$-Hopf algebra. 
The range of $\wedge_\alpha$ is a coideal in $\cH_{LR}$. It is also shown
in  \cite{Zhang} that the $\wedge_\Omega$-Hopf algebra $\cH_{LR}$ is the
free cocycle $\wedge_\Omega$-Hopf algebra, that is, the initial object in
the category of cocycle $\wedge_\Omega$-Hopf algebras.

\smallskip

The external merge operation is modelled on the grafting operations $\wedge_\alpha$
of the Loday-Ronco Hopf algebra, with labels $\alpha\in \{ < , > \}$. The main differences
are that the external merge inverts the order when the first tree is nontrivial,
$$ \cE(T_1,T_2) =T_2 \wedge_> T_1 \ \ \  \text{ for } T_1\neq \bullet  $$
$$ \cE(\bullet ,T_2)=\bullet \wedge_< T_2 $$
and the fact that the operation is only defined on a domain 
${\rm Dom}(\cE)\subset \cH_{ling}\otimes \cH_{ling}$ of pairs of trees 
$(T_1[\beta], T_2[\alpha])$ with matching labels $\beta =\sigma \alpha$.

\smallskip

Both the $\wedge_\alpha$-algebra identity \eqref{veeid} and the cocycle identity \eqref{cocycle}
are still satisfied, whenever all the terms involved are in the domain of the merge operator.
This leads naturally to consider external merge as a structure of {\em partially defined 
cocycle $\wedge_\Omega$-Hopf algebra}. 

\smallskip

Namely, a partially defined cocycle $\wedge_\Omega$-bialgebra 
(or Hopf algebra) is a bialgebra (or Hopf algebra) $(\cH, \Delta, \star)$ together
with a family $\{ \wedge_\alpha \}_{\alpha\in \Omega}$ of binary operations
defined on linear subspaces ${\rm Dom}(\wedge_\alpha)\subset \cH \otimes \cH$,
$$ \wedge_\alpha: {\rm Dom}(\wedge_\alpha)\hookrightarrow \cH \otimes \cH \to \cH $$
that satisfy \eqref{veeid} and  \eqref{cocycle}, whenever all the terms are in 
${\rm Dom}(\wedge_\alpha)$. 

\smallskip

The Hopf algebra $\cH_{ling}$ with the external merge operator $\cE$ is then a
partially defined cocycle $\wedge_\Omega$-Hopf algebra. 

\medskip
\section{New Minimalism, Merge, magmas and Hopf algebras}\label{NewMinSec}

The new version of Minimalism, as presented in  \cite{Chomsky17}, \cite{Chomsky19}, 
\cite{Chomsky23}, and in our mathematical formulation in \cite{MCB}, avoids all the
complications arising in the previous formulation, both coming from the presence of
planar structures and from the need for labeling and projection. By delegating these
aspects to a later externalization procedure, the core computational mechanism of
Merge becomes very transparent and simple from the mathematical perspective
and no longer leads to very different structures underlying Internal and External
Merge. The formulation is also no longer plagued by the problem of partially defined
operations and corresponding domains. 

\smallskip

The whole mathematical structure of Merge in the new version of Minimalism
is discussed in detail in our paper \cite{MCB}, including a proposed model for
externalization. We summarize it here for direct comparison with the
old Minimalism discussed in the previous section.

\smallskip
\subsection{The core computational structure}\label{CoreMergeSec}

In the new version of Minimalism, one starts with a core computational
structure, which is the recursive construction of binary rooted trees, where
trees are now just abstract trees, not endowed with planar structure.
This structure is described mathematically in the following way.

\smallskip

The set $\fT$ of finite binary rooted trees without planar structure is obtained 
as the {\em free non-associative commutative magma} whose elements are
 the balanced bracketed expressions in a single variable $x$, with the
 binary operation (binary set formation)
 \begin{equation}\label{binsetformM}
  (\alpha,\beta)\mapsto \fM(\alpha,\beta)=\{ \alpha, \beta \}\, 
 \end{equation} 
 where $\alpha,\beta$ are two such balanced bracketed expressions.

\smallskip

This description of the generative process of binary rooted trees
is immediate from the identification of these trees with balanced 
bracketed expressions in a single variable $x$, such as
$$ \{ \{ x \{ x x \} \} x \} \longleftrightarrow \Tree [ [ x  [ x x ] ]  x ] \ \ \, . $$
Under this identification, the binary operation of the magma takes two
rooted binary trees and attaches the roots to a new common root
\begin{equation}\label{MergeTrees}
 (T,T') \mapsto \fM(T,T')=\Tree [ T  T' ] \,\,\, . 
\end{equation} 

\smallskip

If one takes the $\Q$-vector space $\cV(\fT)$ spanned by the set $\fT$, the
magma operation $\fM$ on $\fT$ induces on $\cV(\fT)$ the structure of an 
algebra. More precisely, $\cV(\fT)$ is the {\em free commutative non-associative 
algebra generated by a single variable $x$}, or equivalently the {\em free algebra 
over the quadratic operad freely generated by the single
commutative binary operation $\fM$} (see \cite{Holt3}).

\smallskip

One can assign a grading (a weight, measuring the size) to the binary rooted trees 
by the number of leaves, $\ell=\# L(T)$, so that one can decompose the vector
space $\cV(\fT)=\oplus_\ell \cV(\fT)_\ell$ into its graded components (the 
sub-vector-spaces spanned by the trees of weight $\ell$). 

\smallskip

As discussed in \cite{MCB}, this is a very natural and simple mathematical object,
and the generative process for the binary rooted trees can then be seen as the
simplest and most fundamental possible case of recursive solution of a fixed point
problem, of the kind that is known in physics as Dyson--Schwinger equation.

\smallskip

Indeed, one can view the recursive construction of binary rooted trees through
repeated application of the Merge operation \eqref{binsetformM} as the recursive
construction of a solution to the fixed point equation
\begin{equation}\label{fixpt}
X=\fM(X,X) \, ,
\end{equation}
where $X=\sum_\ell X_\ell$ is a formal infinite sum of variables $X_\ell$ in  $\cV(\fT)_\ell$,
with $\ell$ the grading by number of leaves. The problem \eqref{fixpt} can be solved 
recursively by degrees, with $$ X_n=\fM(X,X)_n= \sum_{j=1}^{n-1} \fM(X_j,X_{n-j})\, , $$
with solution $X_1=x$, $X_2=\{ x x \}$, 
$X_3 = \{ x \{ x x \}\} + \{ \{ x x \} x \} = 2 \{ x \{ x x \}\}$, $X_4=2 \{ x \{ x \{ x x \}\} \} +
\{ \{ x x \} \{ x x \} \}$, and so on. The coefficients with which the trees occur in these
solutions count the different possible choices of planar embeddings. 

\smallskip

The equation
\eqref{fixpt} is the simplest case (with the simplest quadratic polynomial) of the
more general form of Dyson--Schwinger equations for rooted trees 
$X =\fB(P(X))$ where $P(X)$ is a polynomial in the formal variable $X$ and
$\fB$ (usually called $B^+$ in the physics literature) is the operation that takes 
a collection of rooted trees $T_1, T_2, \ldots, T_n$, seen as the forest 
$F=T_1 \sqcup T_2 \sqcup \ldots \sqcup T_n$, which appears as  
a monomial in $P(X)$, and constructs a new tree by attaching all the
roots of the $T_i$'s to a single new root (an $n$-ary Merge), 
\begin{equation}\label{fB}
 \fB : T_1 \sqcup T_2 \sqcup \ldots \sqcup T_n  \mapsto \Tree[ $T_1$ $T_2$ $\cdots$ $T_n$ ] \, .
\end{equation}
For a detailed discussion of the role of this operator in physics see, for
instance, \cite{Foissy}, \cite{Yeats}. The case where $P(X)$ has a single quadratic term
gives the binary Merge operation in its core generative process.

\smallskip
\subsection{Syntactic objects}\label{SyntObjSec}

The core computational structure of \S \ref{CoreMergeSec} 
introduces Merge in the most
basic form of binary set formation \eqref{binsetformM}. This
can then be extended to the generative process that gives
rise to {\em syntactic objects}.

\smallskip

One starts with an assigned set $\cS\cO_0$ of {\em lexical items} and
{\em syntactic features} such as $N, V, A$, $P, C, T, D, \ldots$ 

\smallskip

The set $\cS\cO$ of syntactic objects is then identified with the set 
\begin{equation}\label{SOtrees}
\cS\cO  \simeq \fT_{\cS\cO_0}
\end{equation}
of finite binary rooted trees with no assigned planar embedding, and with leaves labelled 
by elements of $\cS\cO_0$. Just as the set of binary rooted trees with no labeling of
leaves has a magma structure with the binary set formation operation $\fM$ of 
 \eqref{binsetformM}, the set of syntactic objects also has a magma structure,
 namely it is the {\em free, non-associative, commutative magma over 
the set $\cS\cO_0$},
\begin{equation}\label{SOeq}
\cS\cO={\rm Magma}_{na,c}(\cS\cO_0, \fM) \, ,
\end{equation}
with the binary Merge operation $\fM$ defined as in \eqref{MergeTrees} for pairs
of trees with labelled leaves. 

\smallskip

Note that the description as elements of the magma ${\rm Magma}_{na,c}(\cS\cO_0, \fM)$
is what is usually referred in the linguistics setting (see  \cite{Chomsky17}, \cite{Chomsky19})
as the description in terms of {\em sets} (in fact {\em multisets}) rather than {\em trees},
since one expresses the elements of the magma in terms of (multi)sets with brackets
corresponding to the Merge operations, rather than representing them in tree form, such as
$$ \{ a, \{ \{ b,c \} , d \} \} \leftrightarrow \Tree [ $a$ [ [ $b$ $c$ ] $d$ ] ] \, , $$
where the tree on the right should not be considered as planar.
As sets $$ \fT_{\cS\cO_0} \simeq {\rm Magma}_{na,c}(\cS\cO_0, \fM) $$ are in bijection, hence
both descriptions are equivalent, but the description as magma is preferred in linguistics
as it emphasizes the generative process. The description in terms of magma elements,
as in the left-hand-side of the example above, also avoids confusion as to whether the
trees have a planar embedding or not: working with (multi)sets rather than with lists
clearly means that no planarity is assumed. We will adopt here the tree notation, just 
because certain mathematical operation we will be using have a simpler and more immediately
visualizable description in 
terms of trees rather than in terms of the corresponding multisets. 

\smallskip

In both the core magma $(\fT,\fM)$ and in the magma \eqref{SOeq} of syntactic
objects, one can introduce a multiplicative unit $1$ satisfying $\fM(T,1)=T=\fM(1,T)$
for all trees, by formally adding a trivial (empty) tree.

\smallskip
\subsection{Workspaces}\label{WorkSpSec}

In the new formulation of Minimalism, the Merge operation acts on workspaces,
consisting of material (lexical items and syntactic objects) available for computation.
The Merge operation updates the workspace for the next step of structure formation.
This notion of workspace is formalized in \cite{CollSta}. Mathematically, as 
discussed in \cite{MCB}, workspaces are just finite disjoint unions of binary rooted trees
(that is, {\em forests}) with leaves labelled by $\cS\cO_0$. Equivalently, workspaces
are multisets of syntactic objects. 

\smallskip

Thus, the set of workspaces can be identified with the set $\fF_{\cS\cO_0}$
of {\em binary rooted forests with no assigned planar structure} (disjoint unions 
of binary rooted trees with no assigned planar structure)
with leaf labels in $\cS\cO_0$. 

\smallskip

Given a workspace $F\in \fF_{\cS\cO_0}$, the material in $F$ that is accessible
for computation consist of the lexical items and all the trees that were obtained
through previous applications of Merge. This inductive definition can be rephrased 
more directly, by defining the {\em set of accessible terms} of $F$.

\smallskip

For a single binary rooted tree $T$ with no assigned planar structure, the
set $Acc(T)$ of accessible terms of $T$ is the set of all subtrees $T_v\subset T$
given by all the descendants of a given non-root vertex of $T$. Indeed, 
these subtrees $T_v$ are exactly all the intermediate trees obtained in 
an iterative construction of $T$ starting from the lexical items or features at the leaves, by repeated
application of the Merge operation \eqref{MergeTrees}. 
We also write $Acc'(T)$ for the set obtained by adding to $Acc(T)$ a copy of $T$
itself, so that
$$ Acc'(T)=\{ T_v \,|\, v \in V_{int}(T) \} \ \ \ \text{ and } \ \ \  Acc'(T)=\{ T_v \,|\, v \in V(T) \}\, , $$
where $V_{int}(T)$ is the set of non-root vertices and $V(T)$ is the set of all vertices of $T$ 
including the root.

\smallskip

For a workspace given by a forest $F=\sqcup_{a\in \cI} T_a \in \fF_{\cS\cO_0}$,
with $T_a$ the component trees and $\cI$ a finite indexing set, the set of accessible
terms {\em of the workspace} is given by
\begin{equation}\label{accF}
Acc(F) = \bigsqcup_{a\in \cI} Acc'(T_a) \, ,
\end{equation}
namely it consists of the syntactic objects, i.e.~the connected components $T_a$ of $F$, 
together with all the accessible terms $\alpha\in Acc(T_a)$ of each component $T_a$. 

\smallskip
\subsection{Merge acting on workspaces}\label{MergeWorkSpSec}

The action of Merge on workspaces is then defined in the new version
of Minimalism as a transformation that produce a new workspace
from a given one, given a choice of a pair of syntactic objects. 

\smallskip

In $S,S' \in \fT_{\cS\cO_0}$ are two given syntactic objects, the
Merge operator $\fM_{S,S'}$ acts on a given workspace $F$ in
the following way. For each of the two arguments of the binary 
operator $\fM_{S,S'}$ the set $Acc(F)$ of accessible terms is
searched for a term matching $S$, respectively $S'$. If 
matching terms are not found the workspace $F=\sqcup_a T_a$ remains the
same. If they are found, say $T_{i,v_i}\simeq S$ and
$T_{j,v_j}\simeq S'$, then the new resulting workspace $F'=\fM_{S,S'}(F) $ is
of the form
\begin{equation}\label{MergeWS}
F' = \fM(T_{i,v_i},T_{j,v_j}) \sqcup T_i/T_{i,v_i} \sqcup T_j/T_{j,v_j} \sqcup \bigsqcup_{a\neq i,j} T_a \, ,
\end{equation}
where the quotients $T_i/T_{i,v_i}$ and $T_j/T_{j,v_j}$ perform the cancellation of the
deeper copies of the accessible terms $T_{i,v_i}$ and $T_{j,v_j}$ in the new workspace. 

\smallskip

The quotient $T/T_v$ of a tree $T$ by an accessible term $T_v\subset T$ is defined here
in the following way, see \cite{MCB}. One first removes the subtree $T_v$ from $T$, leaving
the complement $T\smallsetminus T_v$. There is then a unique maximal binary rooted
tree that can be obtained from this complement $T\smallsetminus T_v$ by contracting
edges, and this resulting binary rooted tree is what we call $T/T_v$. When $v$ is a
non-root vertex of $T$, the quotient $T/T_v$ is equivalently described as obtained
by contracting all of $T_v$ to its root vertex, and also contracting the edge above the 
root vertex of $T_v$ and the other edge out of the vertex above the root of $T_v$.
In the case where $T_v=T$ we have $T/T_v=1$. We also set $T/1=T$.

\smallskip

In this new version of Minimalism, the Hopf algebra structure can be seen as underlying the
action of Merge described as in \eqref{MergeWS}. Indeed, the fundamental
property of the coproduct in a Hopf algebra is providing the list of all the
possible ways of decomposing an object (the term of the algebra the
coproduct is applied to) into a pair of a subobject and the associated
quotient object. In other words, in the case of a tree $T$ one can write a coproduct in the form 
\begin{equation}\label{coprod}
 \Delta(T)=\sum_v F_{\underline{v}} \otimes T/F_{\underline{v}} \, , 
\end{equation} 
where we include the special cases $T\otimes 1$ and $1\otimes T$, and where, for
$\underline{v}=\{ v_i \}_{i=1}^k$ we write $F_{\underline{v}}$ for 
the forest consisting of a union of disjoint subtrees $T_{v_i}\subset T$. 
This can be equivalently described as a forest obtained from an {\em admissible cut} of the tree $T$.
We can view the first terms
\begin{equation}\label{coprodTvpart}
 \Delta_{(2)}(T)=\sum_v T_v \otimes T/T_v \, , 
\end{equation} 
of the coproduct (meaning those where the subforest has a single component) 
as a way of compiling a list of all the accessible terms of $T$ with the
corresponding cancellation of the deeper copy. In these terms, the action \eqref{MergeWS}
can be described as taking, for each of the two arguments of the binary operation $\fM_{S,S'}$
the coproduct $\Delta(F)$ over the entire workspace, which extracts all the accessible terms,
searching among them for matching copies of $S$ and $S'$, merging them if found, and
keeping the other terms of the coproduct that perform the cancellation of the deeper copies.

\smallskip

The $\Q$-vector space $\cV(\fF_{\cS\cO_0})$ spanned by the workspaces,
with the operations of disjoint union $\sqcup$ as product and the coproduct
\eqref{coprod} extended from trees to forests by $\Delta(F)=\sqcup_a \Delta(T_a)$ 
for $F=\sqcup_a T_a$, has the structure of an associative,
commutative, coassociative, non-cocommutative bialgebra 
$(\cV(\fF_{\cS\cO_0}),\sqcup, \Delta)$. The vector space $\cV(\fF_{\cS\cO_0})=\oplus_\ell 
\cV(\fF_{\cS\cO_0})_\ell$ has a grading by number of leaves, compatible with
the operations, so that a coproduct making $\cV(\fF_{\cS\cO_0})$ a Hopf algebra
can be constructed inductively by degrees. 

\smallskip

The action of Merge on workspaces described above is formulated in \cite{MCB}
explicitly in terms of the product and coproduct structure of
$(\cV(\fF_{\cS\cO_0}),\sqcup, \Delta)$, as 
\begin{equation}\label{MergeWSeq}
  \fM_{S,S'} = \sqcup \circ (\fB  \otimes {\rm id}) \circ \delta_{S,S'} \circ \Delta \, ,
 \end{equation}
 where the coproduct $\Delta$ produces the list of accessible terms and corresponding
 cancellations, and the operator $\delta_{S,S'}$ identifies the matching copies.
 These are then fed into Merge by the operation $\fB  \otimes {\rm id}$
 which at the same times produces the new merged tree through the grafting
 operation $\fB$ of \eqref{fB} and performs the cancellation of the deeper
 copies by keeping the corresponding quotient terms produced by the 
 coproduct in the new workspace. The resuling
 new workspace is then produced by taking all these components
 and the new component created by Merge together through the product
 operation $\sqcup$. A more detailed discussion of this operation is
 given in our companion paper \cite{MCB}. 
 
\smallskip
\subsection{Internal and External Merge}\label{IntExtMergeNewSec} 

The action of Merge on workspaces defined by \eqref{MergeWS} and
\eqref{MergeWSeq} contains other forms of Merge in addition to
External and Internal Merge, such Sideward and Countercyclic Merge
(see \S 2.3 of \cite{MCB}). These are not desirable in linguistic terms,
and one expects that a mechanism of Minimal Search would extract
the terms corresponding to External and Internal Merge as the ``least
effort" contributions. Minimal Search refers to the extraction of matching
terms from the list of the accessible terms produced by the coproduct,
where the search in this list is done according to an appropriate
economy principle. 

\smallskip

With the formulation \eqref{MergeWSeq} of the action of Merge on
workspaces, it is shown in \cite{MCB} that the usual form of Minimal Search
can be implemented in this Hopf algebra setting by selecting the leading 
order part of the coproduct with respect to a grading that weights the 
accessible terms $T_v\subset T$ by the 
distance of the vertex $v$ from the root vertex of $T$, so that searching
among terms deeper into the tree becomes less efficient, with respect to
this cost function, than searching near the top of the tree, and taking
the leading order term with respect to this grading has exactly the
same effect as implementing Minimal Search in the way usually
described in the linguistics literature (see \cite{Chomsky17}, \cite{Chomsky19}, 
\cite{Chomsky23}). 

\smallskip

More precisely, one can introduce degrees in the coproduct by taking 
\begin{equation}\label{Deltaepsiloneta}
  \Delta^{(\epsilon,\eta)} : \cV(\fT_{\cS\cO_0}) \to \cV(\fT_{\cS\cO_0})[\epsilon]\otimes_\Q 
 \cV(\fT_{\cS\cO_0})[\eta]\, , 
 \end{equation}
 $$ \Delta^{(\epsilon,\eta)}(T) =\sum_{\underline{v}} \epsilon^{d_{\underline{v}}} \, F_{\underline{v}} \otimes \eta^{d_{\underline{v}}} (T/F_{\underline{v}})\, , $$
 where for $\underline{v}=\{ v_1,\ldots, v_n \}$ a set of vertices $v_i\in V_{int}(T)$, we
 set $d_{\underline{v}}=d_{v_1}+\cdots+d_{v_n}$, with $d_v$ the distance of a vertex $v$ to 
 the root of $T$.
 In the Merge action one correspondingly obtains
\begin{equation}\label{epsMergeWSeq}
\fM^\epsilon_{S,S'} = \sqcup \circ (\fM^\epsilon  \otimes {\rm id}) \circ \delta_{S,S'} 
 \circ \Delta^{(\epsilon,\epsilon^{-1})}
 \end{equation}
 with $\Delta^{(\epsilon,\epsilon^{-1})}$ as in \eqref{Deltaepsiloneta}, and with
 \begin{equation}\label{Mergepsilon}
 \begin{array}{c}
 \fM^\epsilon : \cV(\fT_{\cS\cO_0})[\epsilon,\epsilon^{-1}]\otimes_\Q 
 \cV(\fT_{\cS\cO_0})[\epsilon, \epsilon^{-1}] \to \cV(\fT_{\cS\cO_0})[\epsilon,\epsilon^{-1}] \\[3mm]
 \fM^\epsilon(\epsilon^d \alpha, \epsilon^\ell \beta)=\epsilon^{|d+\ell|}\, \fM(\alpha,\beta)\, .  
  \end{array}
 \end{equation}
It is then shown in \cite{MCB} that taking the leading term (namely the only nonzero term in 
the limit $\epsilon\to 0$) of arbitrary compositions of operators $ \fM^\epsilon_{S,S'}$, one 
finds only (compositions of) Internal and External Merge, while all the remaining
forms of Merge (Sideward, Countercyclic) are of lower order and
disappear in the $\epsilon \to 0$ limit.

\section{Externalization and planarization}\label{ExtLinSec}

In the new formulation of Minimalism, as we discussed above,
Merge happens in the free symmetric form described by the
free commutative non-associative magma  ${\rm Magma}_{na,c}(\cS\cO_0, \fM)$
of \eqref{SOeq} that constructively defines syntactic objects, which are
binary rooted trees with no assignment of planar structure.

\smallskip

In this formulation of Minimalism, the assignment of planar structure to trees
happens {\em after} the action of Merge has taken place, in a further process
of {\em externalization}. 

\smallskip

This is in contrast with older versions of Minimalism (including the case of
Stabler's formulation that we discussed in the previous sections), where
Merge is applied directly on planar trees.

\smallskip

There are suggestions, such as Richard Kayne's LCA (Linear Correspondence Axiom) 
(\cite{Ka1}, \cite{Ka2}), that propose the replacement of externalization with a 
more implicit (and unique) choice of planar embeddings for trees. 
Irrespective of the tenability of this proposal on linguistic ground, we 
discuss in this section some more basic difficulties in its implementation, 
that arise at the formal algebraic level. 

\smallskip

First a comment about terminology:
in the linguistics literature it is customary to use the term {\em linearization} for
the choice of a linear ordering for the leaves of a binary rooted tree. Since
this creates a conflict of terminology with the more common mathematical
use of the word ``linearization", and the choice of a linear ordering of the leaves
is equivalent to the choice of a planar embedding of the tree, we will adopt
here the terminology {\em planarization} (of trees) instead of {\em linearization}
(of the set of leaves). Thus, we will refer to the LCA proposal as a ``planarization''
rather than as a ``linearization algorithm" as usually described. We trust this
will not cause confusion with the readers. 

\smallskip
\subsection{Commutative and non-commutative magmas}\label{MagmaNCsec}

A first important observation is that the Merge operation can happen
{\em before} (as in the new Minimalism) or {\em after} the assignment
of planar structure to trees (as in the old Minimalism), but {\em not}
both consistently. What we mean by this is the following simple mathematical
observation. 

\smallskip

Just as we consider the free commutative non-associative magma  
${\rm Magma}_{na,c}(\cS\cO_0, \fM)$ of the new Minimalism, we
can similarly consider the free non-commutative non-associative magma
$$ \cS\cO^{nc} := {\rm Magma}_{na,nc}(\cS\cO_0, \fM^{nc}) $$
over the same set $\cS\cO_0$. The elements of this magma $\cS\cO^{nc}$
are the {\em planar} binary rooted trees with leaves labelled by
the set $\cS\cO_0$. We write the elements of $\cS\cO^{nc}$ as $T^\pi$,
where $T$ is an abstract (non-planarly embedded) binary rooted tree
and $\pi$ is a planar embedding of $T$.
The non-commutative non-associative magma
operation is given by
$$ \fM^{nc}(T^{\pi_1}_1, T^{\pi_2}_2) = \Tree[ $T_1^{\pi_1}$  $T_2^{\pi_2}$ ] =: T^\pi\, , $$
where now the trees $T^{\pi_1}_1$ and $T^{\pi_2}_2$ are planar and the above tree $T^\pi$ is
assigned the planar embedding $\pi$ where $T_1^{\pi_1}$ is to the left of $T^{\pi_2}_2$.
In particular now $\fM^{nc}(T_1^{\pi_1}, T^{\pi_2}_2)\neq \fM^{nc}(T^{\pi_2}_2, T^{\pi_1}_1)$, 
unlike the case of the commutative $\fM$ of $\cS\cO$.

\smallskip

There is a morphism of magmas $\cS\cO^{nc}\to \cS\cO$ which simply forgets
the planar structure of trees, so that $T^\pi \mapsto T$. It is well defined as a morphism of magmas since
the two different planar trees $\fM^{nc}(T^{\pi_1}_1, T^{\pi_2}_2)$ and $\fM^{nc}(T^{\pi_2}_2, T^{\pi_1}_1)$
have the same underlying abstract (non-planar) tree $\fM(T_1,T_2)$.

\smallskip

On the other hand, there is no morphism of magmas that goes in the opposite direction, from
$\cS\cO$ to $\cS\cO^{nc}$. Indeed, if such morphism existed, then its image would necessarily
be a commutative sub-magma of $\cS\cO^{nc}$, but $\cS\cO^{nc}$ does not contain any 
nontrivial commutative sub-magma. This can be seen easily as, if a tree $T^\pi$ is
contained in a commutative sub-magma of $\cS\cO^{nc}$, then $\fM^{nc}(T^\pi, T^\pi)$
also is, but $\fM^{nc}(\fM^{nc}(T^\pi, T^\pi), T^\pi))\neq \fM^{nc}(T^\pi, \fM^{nc}(T^\pi, T^\pi))$
contradicting the fact that the sub-magma is commutative. 

\smallskip

This fact has the immediate consequence that we stated above, namely that if a
free symmetric Merge takes place {\em before} any assignment of planar structure to
trees, then it cannot also consistently apply {\em after} a choice of planarization.
In other words, if $\Sigma$ is any choice of a section of the projection $\cS\cO^{nc}\to \cS\cO$
(that is, an assignment of planarization) then one {\em cannot have} compatible Merge
operations satisfying $\fM^{nc}(\Sigma(T_1),\Sigma(T_2))=\Sigma(\fM(T_1,T_2))$. Merge can act on
abstract trees as in the new Minimalism or on planar trees as in the old Minimalism,
but these two view are mutually exclusive. 

\smallskip

Descriptions of Merge at the level of planar trees, as in the old Minimalism,
involve a partially defined operations with restrictions on domains, based on some label
assignments. We have discussed this explicitly in the case of Stabler's formulation
but analogous conditions exist in other formulations of the old Minimalism. In view of
the considerations above, this can be read as evidence of the fact that 
one is trying to describe at the level of planar trees a Merge that is really taking
place at the underlying level of non-planar abstract trees, correcting for the
incompatibility described above by restricting the domain of applicability. 
In particular, this means that the significant complications in the underlying
algebraic structure of External and Internal Merge that we have observed in
the case of Stabler's formulation, similarly apply to any other formulation
that places Merge after the planar embedding of trees. 

\smallskip

This issue does not arise within the
formulation of the new Minimalism, since the planarization that happens in externalization
is simply a non-canonical (meaning dependent on syntactic parameters) choice of a section
$\Sigma$ of the projection $\cS\cO^{nc}\to \cS\cO$. The only requirement on $\Sigma$ is to be compatible
with syntactic parameters, not to be a morphism with respect to the magma operations
$\fM$ and $\fM^{nc}$, since in this model Merge only acts before externalization as $\fM$
(not after externalization as $\fM^{nc}$). In our previous paper \cite{MCB} we
describe externalization in the form of a correspondence. This represents the
two step procedure that first choses a non-canonical section $\Sigma$ and then quotients
the image by eliminating those planar trees obtained in this way that are not
compatible with further (language specific) syntactic constraints. Merge is not
applied anywhere in this externalization process, which happens {\em after} the
results of Merge have been computed at the level of trees without planar structure.

\smallskip
\subsection{Planarization versus Externalization}\label{LCAsec}

Proposals such as Kayne's LCA planarization of trees, \cite{Ka1}, \cite{Ka2} (see
also Chapter~7 of \cite{Horn} and \cite{LaCa} for a short summary), 
suggest the replacement of externalization with a different way of constructing
planarization. This relies on the idea that the abstract trees are endowed with
additional data (related to heads, maximal projection, and c-command relation) 
that permits a {\em canonical} choice of planar structure. We discuss two
different mathematical difficulties inherent in this proposal.

\smallskip

First let's assume that indeed additional data on the abstract syntactic tree
suffice to endow them with a unique canonical choice of linearization. (We will
discuss the difficulties with this assumptions below.) In this case, we would
have a map, which we call $\Sigma^{LCA}$ that assigns to an abstract binary rooted
tree $T$ a corresponding, uniquely defined non-planar tree $\Sigma^{LCA}(T)=T^{\pi_{LCA}}$,
with $\pi_{LCA}$ the linear ordering constructed by the LCA algorithm. 

\smallskip

By our previous observations on the morphisms of magmas, we will necessarily have in general that 
$$ \fM^{nc}(\Sigma^{LCA}(T_1), \Sigma^{LCA}(T_2)) \neq \Sigma^{LCA}( \fM(T_1,T_2))\, , $$
so that $\Sigma^{LCA}$ {\em cannot be compatible} with asymmetric Merge of planar trees. 

\smallskip

The only way to make it compatible with Merge would be to define Merge on the
image of $\Sigma^{LCA}$ not as the asymmetric Merge of planar trees but as
$$ \fM^{LCA}(\Sigma^{LCA}(T_1), \Sigma^{LCA}(T_2)):= \Sigma^{LCA}(\fM(T_1,T_2)) \, . $$
This, however, simply creates an isomorphic copy of the commutative
magma ${\rm Magma}_{na,c}(\cS\cO_0, \fM)$ (by a choice of a 
particular representative in each equivalence class of the projection
$\cS\cO^{nc}\to \cS\cO$). This would imply that application of the planarization
$\Sigma^{LCA}$ has no effect, in terms of the properties of Merge, with respect to working
directly with the free symmetric Merge on abstract non-planar trees. 
The alternative is, as in externalization, to not require any compatibility between
$\Sigma^{LCA}$ and Merge: in other word, even if LCA replaces externalization,
Merge is still only happening in the form of free symmetric Merge, 
at the level of the abstract non-planar trees, and not after planarization. 

\smallskip

We now look more specifically at the proposals for how the planarization $\Sigma^{LCA}$
should be obtained, to highlight a different kind of difficulty. 

\smallskip

In an (abstract) binary rooted tree $T$, two vertices $v_1,v_2$ are 
sisters if  there is a vertex $v$ of $T$ connected to both $v_1$ and $v_2$.
A vertex $v$ of $T$ dominates another vertex $w$ if $v$ is 
on the unique path from the root of $T$ to $w$. 
A vertex $v$ in $T$ c-commands another vertex $w$ if
neither dominates the other and the lowest vertex that 
dominates $v$ also dominates $w$.  A vertex $v$ asymmetrically
c-commands a vertex $w$ if $v$ c-commands $w$ and $v,w$ are
not sisters. Asymmetric c-command defines a partial ordering of
the leaves of $T$.  In order to extend this partial ordering relation,
instead of using directly the asymmetric c-command relation to
define the order structure, one uses maximal projections. 
Namely, one requires that a leaf $\ell$ precedes another leaf $\ell'$
if and only if either $\ell$ asymmetrically c-commands $\ell'$ or
a maximal projection dominating $\ell$ c-commands $\ell'$.

\smallskip

Even with this extension using maximal projections, there are
issues in making this a total ordering, as discussed for instance
in Chapter~7 of \cite{Horn} and in \cite{LaCa}. 

\smallskip

Since we have been discussing in the previous sections the
Stabler formalism, it is easy to see this same issue in terms of
the labels $>$ and $<$ assigned to the internal vertices of
(planar) trees in Stabler's formulation. The difference is that
in Stabler one starts with a tree that already has a planar
assignment and uses heads, maximal projection, and c-command
to obtain a new planar projection. In that case, as we discussed
already, the labels $>$ and $<$ are assigned to the root and the 
non-leaf vertices of a planar tree by pointing, at each vertex $v$, in
the direction of the branch where the head of the tree $T_v$ resides.
If this assignment of labels is well defined, then the new planar
structure of the tree can be obtained simply by flipping
subtrees about their root vertex every time the vertex is labelled
$<$, until all the resulting vertices become labelled by $>$.

\smallskip

The implicit assumption that makes this possible
is that every subtree $T_v$ of a given tree $T$ has a head,
that is, a marked leaf. This assumption on heads is necessary both for
the labeling by $>$ and $<$ in Stabler's formalism and for
the use of maximal projections for the definition of ordering in LCA,
since a maximal projection is a subtree $T_v$ of $T$ that is not
strictly contained in any larger $T_w$ with the same head. 

\smallskip

In the case of the labels $>$ and $<$, the problem of what
label is assigned to the new root vertex, when External or 
Internal Merge is performed, requires restrictions on
the domain of applicability of these Merge operations, 
depending on conditions on the labels at the heads of the
trees used in the Merge operation (which makes them
partially defined operations). It also requires other 
further steps, such as allowing 
Merge results to be unlabeled during derivation and labelled 
at the end so that successive-cyclic raising can remove 
the elements responsible for un-labelability. (This refers
to the same linguistic problem with Stabler's formalism
that we discussed above, at the end of \S \ref{OldIntMergeSec},
mentioned to us by Riny Huijbregts).

\smallskip

In the case of LCA, the problem again arises from the
fact that a result of Merge need not have the heads of
the two merged trees in an asymmetric c-command relation.
This implies that, even with the introduction of heads and
maximal projections, the assumption that all trees have
a head marked leaf cannot always be satisfied, hence
one cannot obtain a total ordering
of the leaves (a unique choice of planar embedding).

\smallskip

Partial corrections to this problem in LCA are suggested
using movement (see \cite{Horn} p.230-231), which is similar 
in nature to the point mentioned above regarding un-labelable 
structures in Stabler's formalism, or by introducing ``null heads"
in the structure, or by morphological reanalysis that hides certain 
items from LCA. In any case, the fundamental difficulty in constructing 
a planarization algorithm based on heads and maximal projections 
can be summarized as follows.

\smallskip

We define a {\em head function} on an (abstract) binary rooted tree $T$
as a function $h_T: V^o(T)\to L(T)$ from the set $V^o(T)$
of non-leaf vertices of $T$ to the set $L(T)$ of leaves of $T$,
with the property that if $T_v \subseteq T_w$ and
$h_T(w)\in L(T_v)\subseteq L(T_w)$, then $h_T(w)=h_T(v)$.
We write $h(T)$ for the value of $h_T$ at the root of $T$.
This general definition is designed to abstract the properties of the head in
the syntactic sense. 

\smallskip

Consider then pairs $(T, h_T)$ and $(T', h_{T'})$ of trees with given head functions.
There are exactly two choices of a head function on the 
Merge $\fM(T,T')$, corresponding to whether $h(T)$ or $h(T')$ 
is equal to $h(\fM(T,T'))$. Since the trees $T,T'$ are not planar and 
$\fM$ is the symmetric Merge, there is no consistent way of making
one rather than the other choice of $h_{\fM(T,T')}$, at each 
application of $\fM$. This implies that, on a given binary rooted tree $T$
there are $2^{\# V^o(T)}$ possible head functions. We can think of
any such choice as the assignment, at each vertex $v\in V^o(T)$ of a
marking to either one or the other of the two edges exiting $v$ in the 
direction away from the root. 

\smallskip

By thinking of $h_T$ as an assignment of a marking to one 
of the two edges below each vertex, the head function $h_T$ 
determines a planar embedding of $T$
by putting under each vertex the marked edge to the left. 
Thus, the problem of constructing planar embeddings of
abstract binary rooted trees can be transformed into the
problem of constructing head functions. 

\smallskip

The LCA algorithm aims at obtaining a special assignment
$T \mapsto h_T$ of head functions $h_T$ to abstract binary 
rooted trees $T$ that is somehow determined uniquely by
the properties of the labeling set $\cS\cO_0$ of the
leaves of the trees. Let us denote by $\lambda(\ell)$ the
label in $\cS\cO_0$ assigned to the leaf $\ell\in L(T)$.

\smallskip

This is where the main difficulty arises. For instance, 
if the labeling set $\cS\cO_0$ happens to be a totally
ordered set (which is not a realistic linguistic assumption),
as long at two trees $(T, h_T)$ and $(T', h_{T'})$
have head functions with labels $\lambda(h(T))\neq \lambda(h(T'))$,
there is always a preferred choice of head function  on $\fM(T,T')$,
which is the one in which the two subtrees $T$ and $T'$ are
ordered according to the ordering of the labels $\lambda(h(T))$
and $\lambda(h(T'))$ in $\cS\cO_0$. However, this excludes the
case where the leaves $h(T)$ and $h(T')$ may have the
same label in $\cS\cO_0$. So even under the unrealistically
strong assumption that labels are taken from a totally ordered
set, a planarization algorithm based on the construction of a
head function cannot be defined on all the syntactic objects
in $\cS\cO$. 

\smallskip

At the level of the underlying algebraic structure, the
issue with planarization is therefore twofold. It does not
provide an alternative to Merge acting on the non-planar
trees, for the reasons mentioned above on maps of
magmas. At the same time, a choice of planar
embedding that is independent of syntactic parameters
and is based only on heads of trees would require a
canonical construction of head functions from properties
of the labeling set $\cS\cO_0$, but this cannot be done
consistently on the entire set of syntactic objects $\cS\cO$
produced by the free symmetric Merge $\fM$.

\smallskip
\section{Conclusions}

We have seen from this comparative analysis of the algebraic structures
underlying one of the older versions of Minimalism (Stabler's Computational
Minimalism) and Chomsky's newer version of Minimalism that the 
new version has a simpler mathematical structure with a unifying 
description of Internal and External Merge, and with a core generative
process that reflects the most fundamental magma of binary set formation,
generating the binary rooted trees without planar structure. 

\smallskip

The more complicated mathematical structure of Stabler's Minimalism
is caused by several factors. The intrinsic asymmetry of the Internal
Merge is due to working with {\em planar} binary rooted trees. Abandoning
the idea that planar embeddings should be part of the core
computational structure of Minimalism is justified linguistically by
the relevance of structures (abstract binary rooted trees) rather
than strings (linearly ordered sets of leaves, or equivalently planar
embeddings of trees) in syntactic parsing, see \cite{Eve}. Thus,
working with abstract trees without planar embeddings is one of
the simplifying factors of the new Minimalism. The other main
issue that complicates the mathematical structure of the older
versions of Minimalism is the very different nature of the Internal
and External Merge operations: in the case of Stabler's Minimalism
analyzed here these two forms of Merge relate to two very different algebraic objects
(operated algebras and right-ideal coideals) hence they cannot be
reconciled as coming from the same operation, while in the new
Minimalism both Internal and External Merge are cases of the
same operation, and both arise as the leading terms with
respect to the appropriate formulation of Minimal Search. Finally, another
main issue that makes the mathematical structure of older
versions of Minimalism significantly more complicated is the
presence of conditions on labels that need to be matched for Internal and
External Merge to be applicable, related to the problem
of projections discussed in \cite{Chomsky13}. Mathematically
this makes all operations only partially defined on particular
domains where conditions on labels are met. As we discussed,
this creates problems with having to work with partially
defined versions of various algebraic structures and it
significantly complicated iterations of the Merge operations,
where the conditions on domains compound. Since the
conditions on labels are absent from the fundamental structure
of the new Minimalism, this problem of dealing with
partially defined operations also disappears, leading to
another simplification at the level of the mathematical
structures involved. 

\smallskip

Within the new formulation of Minimalism, the planar structure
of trees is introduced as a later step of externalization, not at
the level of the Merge action. The choice of planar structure in
externalization is done through a non-canonical (that is,
dependent on syntactic parameters) section of the projection
from planar to abstract trees. We show that proposed construction
of a unique canonical choice of planar embeddings, based on
heads of trees, maximal projections, and c-command, run into
difficulties at the level of the underlying algebraic structure.

\medskip
\subsection*{Acknowledgments} 
We thank Riny Huijbregts for helpful comments and feedback, and
Richard Kayne for useful discussions. 
The first author acknowledges support from NSF grant DMS-2104330 and 
FQXi grants FQXi-RFP-1 804 and FQXi-RFP-CPW-2014, SVCF grant 2020-224047, 
and support from the Center for Evolutionary Science at Caltech.

\end{document}